\documentclass[conference]{IEEEtran}
\IEEEoverridecommandlockouts
\usepackage{cite}
\usepackage{amsmath,amssymb,amsfonts}
\usepackage{graphicx}
\usepackage{textcomp}
\usepackage{xcolor}
\usepackage{amsmath}
\usepackage{booktabs}
\usepackage{colortbl}
\usepackage{algpseudocode}
\usepackage{algorithm} 
\usepackage{adjustbox}
\usepackage{color}
\usepackage{amsfonts}
\usepackage{amssymb}
\usepackage{bbding}
\usepackage{multirow}
\usepackage{graphicx}
\usepackage{amsmath}
\usepackage{booktabs}
\usepackage{colortbl}
\usepackage{algpseudocode}
\usepackage{algorithm} 
\usepackage{adjustbox}
\usepackage{color}
\usepackage{amsfonts}
\usepackage{amssymb}
\usepackage{bbding}
\usepackage{multirow}
\usepackage{rotating}
\def\BibTeX{{\rm B\kern-.05em{\sc i\kern-.025em b}\kern-.08em
    T\kern-.1667em\lower.7ex\hbox{E}\kern-.125emX}}

\makeatletter
\newcommand{\linebreakand}{%
  \end{@IEEEauthorhalign}
  \hfill\mbox{}\par
  \mbox{}\hfill\begin{@IEEEauthorhalign}
}
\makeatother

\begin{document}

\title{LSAS: Lightweight Sub-attention Strategy for Alleviating Attention Bias Problem*
\thanks{This work was supported in part by National Natural Science Foundation of China (NSFC) under Grant No.62206314 and Grant No.U1711264, GuangDong Basic and Applied Basic Research Foundation under Grant No.2022A1515011835, China Postdoctoral Science Foundation funded project under Grant No.2021M703687. \\ Corresponding Author: Jinghui Qin.}
}

\author{
\IEEEauthorblockN{1\textsuperscript{st} Shanshan Zhong}
\IEEEauthorblockA{\textit{Sun Yat-sen University} \\
Guangzhou, China \\
zhongshsh5@mail2.sysu.edu.cn}

\and
\IEEEauthorblockN{2\textsuperscript{nd} Wushao Wen}
\IEEEauthorblockA{\textit{Sun Yat-sen University} \\
Guangzhou, China \\
wenwsh@mail.sysu.edu.cn}

\and
\IEEEauthorblockN{3\textsuperscript{rd} Jinghui Qin}
\IEEEauthorblockA{\textit{Guangdong University of Technology} \\
Guangzhou, China \\
scape1989@gmail.com}

\linebreakand 
\IEEEauthorblockN{4\textsuperscript{th} Qiangpu Chen}
\IEEEauthorblockA{\textit{Sun Yat-sen University} \\
Guangzhou, China \\
chenqp8@mail2.sysu.edu.cn}

\and
\IEEEauthorblockN{5\textsuperscript{th} Zhongzhan Huang}
\IEEEauthorblockA{\textit{Sun Yat-sen University} \\
Guangzhou, China \\
huangzhzh23@mail2.sysu.edu.cn}
}

\maketitle

\begin{abstract}
In computer vision, the performance of deep neural networks (DNNs) is highly related to the feature extraction ability, i.e., the ability to recognize and focus on key pixel regions in an image. However, in this paper, we quantitatively and statistically illustrate that DNNs have a serious attention bias problem on many samples from some popular datasets: (1) Position bias: DNNs fully focus on label-independent regions; (2) Range bias: The focused regions from DNN are not completely contained in the ideal region. Moreover, we find that the existing self-attention modules can alleviate these biases to a certain extent, but the biases are still non-negligible. To further mitigate them, we propose a lightweight sub-attention strategy (LSAS), which utilizes high-order sub-attention modules to improve the original self-attention modules. The effectiveness of LSAS is demonstrated by extensive experiments on widely-used benchmark datasets and popular attention networks. We release our code to help other researchers to reproduce the results of LSAS~\footnote{https://github.com/Qrange-group/LSAS}. 
\end{abstract}


\section{Introduction}
\label{sec:intro}

\begin{figure}[t]
  \centering
  \includegraphics[width=\linewidth]{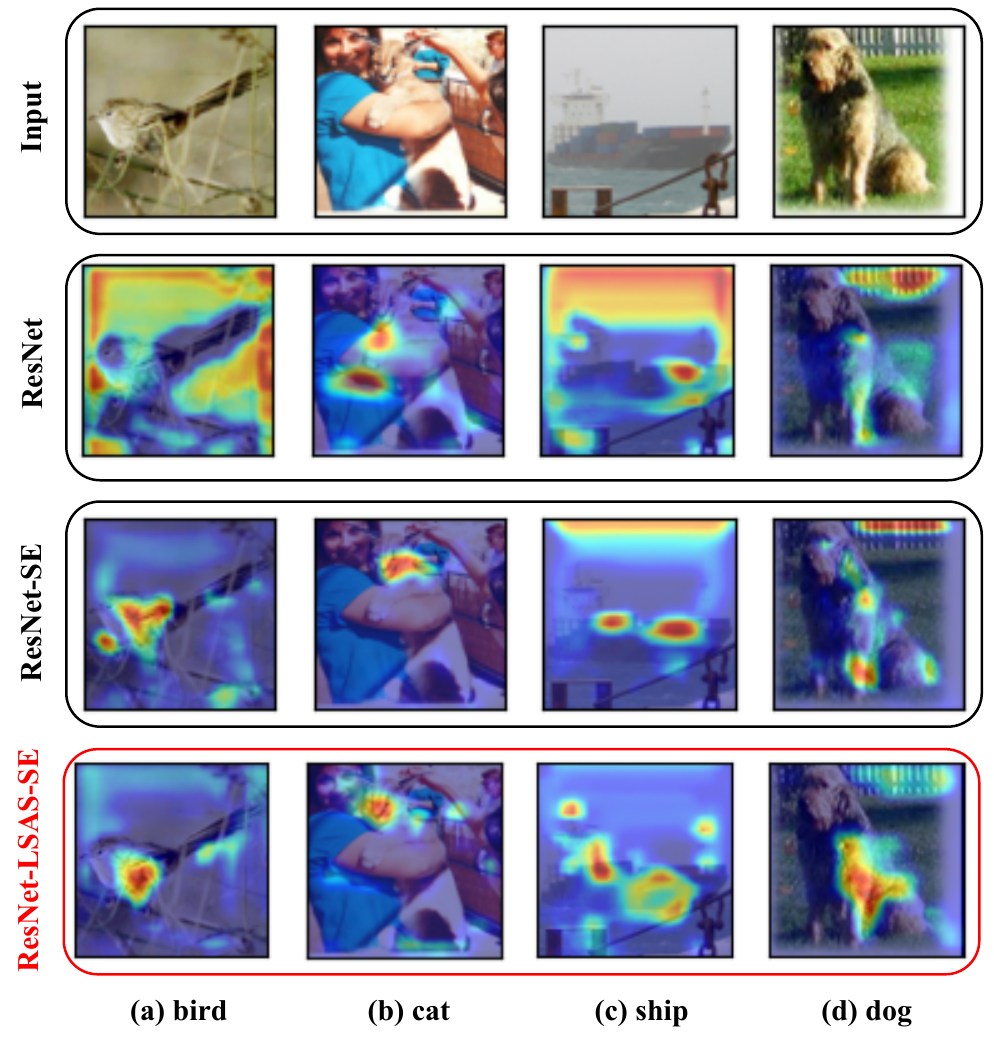}
  \caption{The visualization of model feature maps based on Grad-CAM, with the STL10 dataset and ResNet164 as the experimental setup. The Grad-CAM technique highlights the regions of the image that contribute most significantly to the model's decision-making process. }
\label{fig:section1}
\end{figure}

Deep neural networks (DNNs) have been empirically confirmed to have efficient and reliable feature extraction capabilities which play a fundamental role in the performance of DNNs~\cite{liang2020instance,kuang2021face} through comprehensive experimental results under various tasks~\cite{he2016deep,zeiler2014visualizing,mcnamara2017developing}. 
Specifically, the feature extraction ability of DNNs is mainly reflected in whether it can recognize and pay attention to key pixel regions in an image~\cite{zhu2021residual,guo2021ssan} in computer vision.
As depicted in Fig.~\ref{fig:section1}, a popular interpretability technology, i.e., Grad-CAM~\cite{2017Grad}, is adopted to explicitly visualize the regions where DNNs attend in the form of heat maps. 
From the results, we can find that although the vanilla ResNet~\cite{he2016deep} achieves good performance, there are non-negligible attention bias problems in key semantic feature extraction:
(1)~\textbf{Position bias.} In the examples illustrated in Fig.~\ref{fig:section1}(a)(b), ResNet only attends to the label-independent background region rather than the region of the bird and the cat. 
These position biases can make the features extracted by DNNs sensitive to background information, resulting in error predictions. 
(2)~\textbf{Range bias.} As shown in Fig.~\ref{fig:section1}(c)(d), ResNet is unable to attend to the overlay region of the label while attending to some extra regions such as sky and fence.

\begin{figure*}[t]
  \centering
  \includegraphics[width=\linewidth]{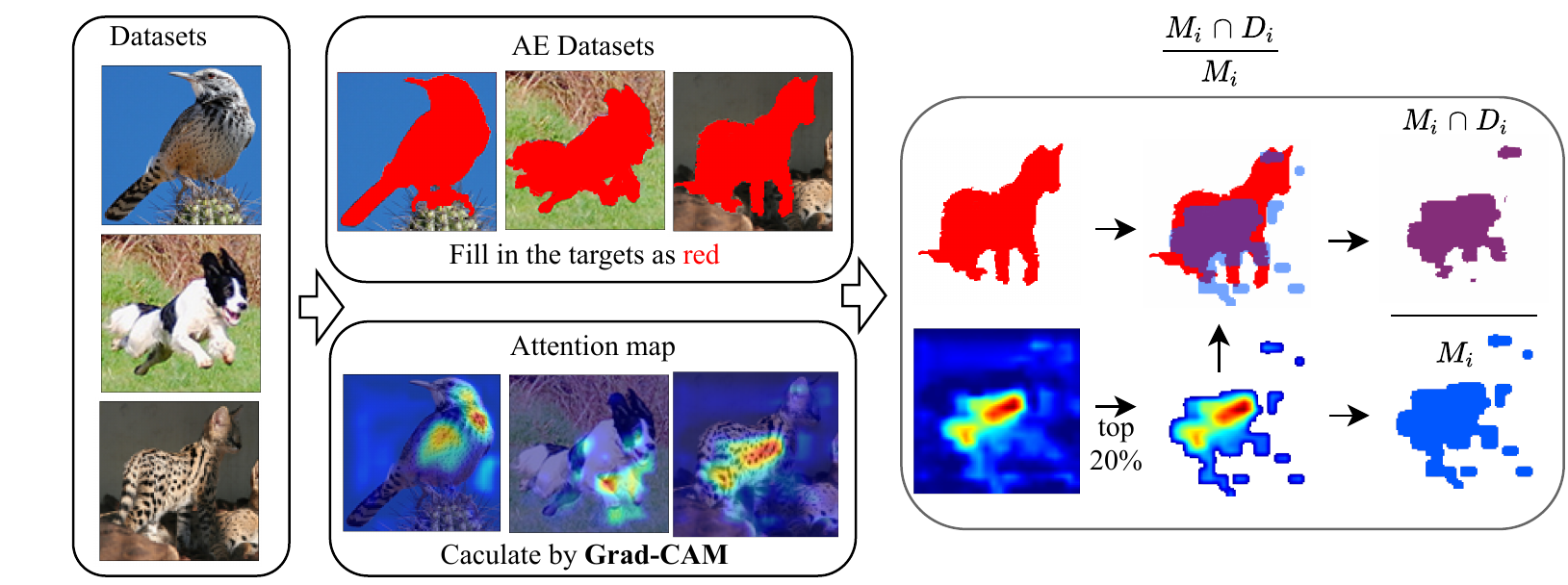}
  \caption{The process of AE calculation, where the red region denotes the ideal focused region. The mathematical symbols in the figure correspond to Eq.~(\ref{eq:ae}). Specifically, $D_i$ represents the ideal focused region on the $i$-th image from dataset $D$, while $M_i$ denotes the focused region generated by the DNNs. }
\label{fig:ae}
\end{figure*}

We also find that these biases can be effectively mitigated by self-attention mechanism~\cite{huang2020dianet,hu2018squeeze} that can focus on important information~\cite{anderson2005cognitive}.
For example, the classic self-attention module SE~\cite{hu2018squeeze} makes ResNet refocus on the regions of bird and cat respectively as shown in Fig.~\ref{fig:section1}(a)(b). However, in Fig.~\ref{fig:section1}(c)(d), although ResNet-SE can focus on less irrelevant regions than ResNet, it still pays attention to other label-independent regions like the sky and the fence.
Therefore, there is still substantial room for improving attention mechanism to focus on the target region as much as possible while ignoring irrelevant regions. 

In Section~\ref{sec:ana}, we further confirm these two biases statistically through a quantitative index and dataset, called Attention Efficiency (AE) and AE-dataset (including AE-STL10 and AE-ImageNet10), respectively. In AE-dataset, we annotated the ideal region of the corresponding label in each image.
These quantitative results of AE reveal that the self-attention mechanism can indeed mitigate attention bias, but there is still a gap between the focused region and ideal region from each example of AE-dataset, which limits the performance of DNNs.
Inspired by the debiasing effect of the self-attention mechanism on DNNs, we propose a Lightweight Sub-attention Strategy (LSAS), which considers utilizing the self-attention mechanism to help the original self-attention mechanism further debias the attention bias problem.
In Fig.~\ref{fig:section1} (a)(b), the attention regions of LSAS-SENet and SENet are similar. In Fig.~\ref{fig:section1} (c)(d), the attention regions of LSAS-SENet are more concentrated on the target, i.e., the ship and the dog, and the attention to irrelevant regions such as the sky and fence are reduced compared to SENet's visualization. These visualization results once again demonstrate the significant effect of LSAS in mitigating attention bias.
Our main contributions are as follows:
\begin{itemize}
     \item We propose AE and AE-datasets to quantitatively confirm that many existing self-attention modules still have a non-negligible attention bias problem, which has a negative effect on the performance of DNNs.
     \item We propose LSAS to alleviate the attention bias of the original self-attention mechanism. Experiments on multiple self-attention modules show that LSAS can improve DNN performance significantly while reducing the parameter amount and increasing computation speed. 
\end{itemize}

\section{Attention Bias Problem Analysis}
\label{sec:ana}

In section~\ref{sec:intro}, we demonstrate that DNNs exhibit a non-negligible attention bias, which can be partially alleviated through self-attention mechanisms. In this section, we propose the Attention Efficiency (AE) and AE-dataset to quantitatively measure the debiasing ability of DNNs. For AE-dataset, we randomly sample 12 images from each of 10 classes in STL10 and ImageNet and fill the ideal focused region as red to form AE-STL10 and AE-ImageNet10 respectively as shown in Fig.\ref{fig:ae}. 
\begin{equation}
\begin{aligned}
\mathbf{AES}(M_i, D_i) &= \begin{cases}                   
1, & \text{if } \frac{M_i \cap D_i}{M_i} > \lambda  \\
0, & \text{others},  
\end{cases} \\
\mathbf{AE}(M, D) &= \frac{1}{|D|} \sum^{|D|}_i \mathbf{AES}(M_i, D_i),
\end{aligned}
\label{eq:ae}
\end{equation}
For DNN $M$ and AE-dataset $D$, we calculate AE via Eq.~(\ref{eq:ae}), where $D_i$ represents the ideal focused region on the $i$-th image from $D$. We treat the region consisted of the area which has the top 20\% of attention values measured by Grad-CAM as the focused region $M_i$ from a DNN.

\begin{table}[htbp]
  \centering
  \caption{AE (\%) of ResNet with different depths and with different attention modules on AE-STL10 and AE-Imagenet10. \textbf{Org} represents vanilla ResNet. }
  \resizebox*{\linewidth}{!}{
    \begin{tabular}{lcccccc}
    \toprule
    \multirow{2}[4]{*}{Dataset} & \multirow{2}[4]{*}{Model} & \multicolumn{5}{c}{ResNet} \\
\cmidrule{3-7}          &       & Org      & SENet~\cite{hu2018squeeze}    & LSAS-SENet & CBAM~\cite{woo2018cbam}  & LSAS-CBAM \\
    \midrule
    AE-STL10 & ResNet164   & 9.17  & 22.50 & 38.33 & 30.00 & 31.67 \\
    AE-ImageNet10 & ResNet50    & 26.67 & 34.17 & 40.83 & 29.17 & 38.33 \\
    \bottomrule
    \end{tabular}%
  }
  \label{tab:AE-example}%
\end{table}%

\begin{figure*}[htp]
  \centering
  \includegraphics[width=\linewidth]{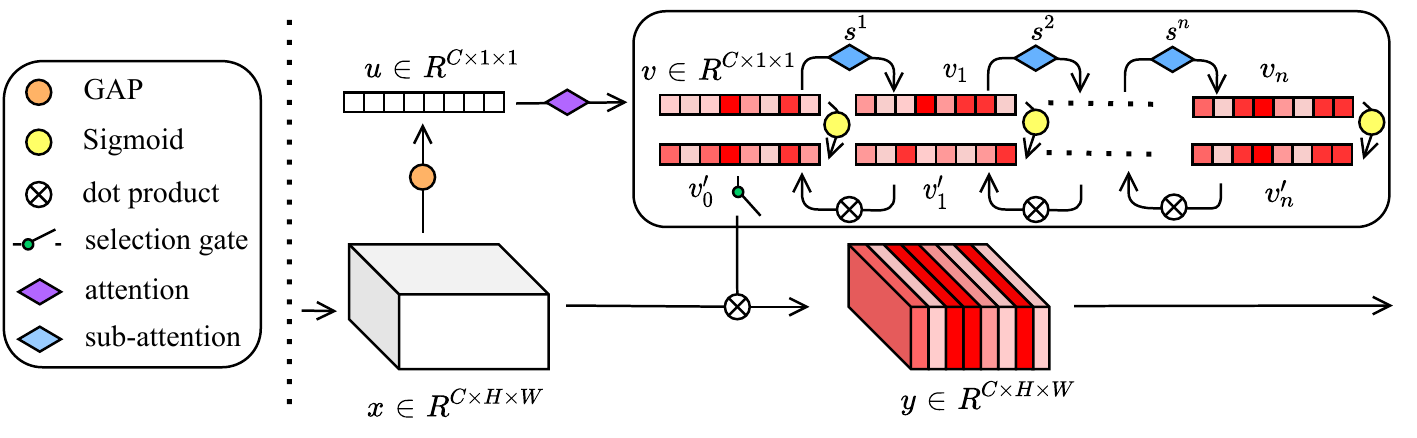}
  \caption{The structure of an attention block, where the box content is the illustration of LSAS. }
\label{fig:LSAS}
\end{figure*}

$\mathbf{AES}(M_i, D_i)$ denotes the AE score for $M$ on the $i$-th image, which is the ratio of $D_i\cap M_i$ and $M_i$, and $\mathbf{AE}(M, D)$ denotes the AE score for $M$ in $D$. We set the threshold $\lambda \in [0, 1]$ as 0.8 in this paper. If the $\mathbf{AE}(M, D)$ is large, the DNN the focused region from the DNN is consistent with the ideal focused region, which means the model has powerful feature extraction ability. Otherwise, the DNN may tend to focus on the wrong region, leading to poor prediction.
As the result about AE shown in Table~\ref{tab:AE-example}, the popular self-attention modules, SENet and CBAM, can mitigate the attention bias of ResNet but the bias still can not be ignored.
Our proposed method LSAS can improve the AE of SENet by 70.36\% and 19.49\% on AE-STL10 and AE-ImageNet10 respectively, and improve the AE of CBAM by 5.57\% and 31.58\% on AE-STL10 and AE-ImageNet10 respectively.

\section{Lightweight Sub-attention Strategy}

In this section, we first review the structure of the original self-attention modules and then we illustrate the design of LSAS.
 
For the original self-attention modules, the input feature of the module is $x \in R^{C \times H \times W}$, where $C$, $H$, and $W$ stand for the number of channels, height, and width of the feature respectively. 
Global average pooling (GAP) is used to produce a global information embedding $u = \mathbf{GAP}(x)$, where $u \in R^{C \times 1 \times 1}$. 
The attention module $g(\cdot)$ is applied to extract the attention feature $v = g(u)$, where $v \in R^{C \times 1 \times 1}$. 
We obtain the attention-debiased output $y = x \otimes \sigma (v)$, where $\otimes$ denotes element-wise multiplication and $\sigma(\cdot)$ is the Sigmoid function.

Motivated by the fact that the self-attention mechanism can mitigate the attention bias, we further propose a multi-order sub-attention strategy LSAS to alleviate the attention bias and improve the performance of DNNs. In other words, we use the self-attention mechanism to enhance the self-attention mechanism.
As shown in Fig.~\ref{fig:LSAS}, setting $n \in [0, \infty)$ be the order of LSAS, the forward stage of the $i$-th sub-attention is as follows:

\begin{equation}
\begin{aligned}
v_{i} = s(v_{i-1})^i = v_{i-1} \otimes \gamma_i + \beta_i, 
\end{aligned}
\label{eq:dsa}
\end{equation}
where $i \in [1, \infty)$, $s(\cdot)^i$ is the $i$-th sub-attention module, and $v_{i-1}$ represents the input of the $i$-th sub-attention. 
Specifically, $v_0 = v$ and $s(v_0)^0 = v_0$. 
A pair of learnable parameters $\gamma_i$, $\beta_i$ are used to scale and shift $v_{i-1}$ to refine the sub-attention map $v_i$.

We obtain the value of each order of sub-attention modules through the forward stage shown in Eq.~(\ref{eq:dsa}), and then in the backward stage, the output of each order of sub-attention modules $v_{i}^{'} = \sigma (v_{i}) \otimes v_{i+1}^{'}$.
Specifically, for the $n$-th sub-attention module, $v_{n}^{'} = \sigma (v_n)$ and if n = 0, $v_0^{'} = \sigma (v_0) = \sigma (v)$. 
Therefore, in attention modules enhanced by multi-order sub-attention modules, we obtain the attention-debiased output $y = x \otimes v_0^{'}$.

\begin{algorithm}[t]  
\small
    \caption{The algorithm of producing attention map from attention modules enhanced by LSAS}
    \textbf{Input:}  A feature map $x\in R^{C\times H\times W}$; attention operator $g$ and $n$ pair of learnable parameters. 
    
    \textbf{Output:} The attention-debiased output $y$.
        
    \begin{algorithmic}[1]

    \State Calculate $u \gets \mathbf{GAP}(x)$;   \algorithmiccomment{Global pooling module}
    
    \State Calculate $v_0 \gets \sigma (g(u))$;                \algorithmiccomment{Attention module} 
       
    \State\algorithmiccomment{Sub-attention module} 

    \State{\textbf{for} $i$ \textbf{from} $1$ \textbf{to} $n$ \textbf{do}}     \algorithmiccomment{(a) Forward stage} 
	\State{\indent Calculate $v_i$ by Eq.~(\ref{eq:dsa})};
	\State {\textbf{end}}
	
	\State{\textbf{for} $i$ \textbf{from} $(n-1)$ \textbf{to} $0$ \textbf{do}}  \algorithmiccomment{(b) Backward stage} 
	\State{\indent Calculate $v_{i}^{'} \leftarrow \sigma (v_{i}) \otimes v_{i+1}^{'}$};
	\State {\textbf{end}}
    \State $\mathbf{SG} (\cdot) \gets $ Eq.~(\ref{eq:select});   \algorithmiccomment{(c) Selection gate} 
    \State Calculate $y \gets x \otimes \mathbf{SG}(v_0^{'})$;  
    \State\Return $y$
    \end{algorithmic} 
    \label{alg:ean}   
\end{algorithm}

\begin{table*}[h]
  \centering
  \caption{Testing accuracy (\%) and Frames Per Second (FPS) on CIFAR10, CIFAR100, and STL10. \#P(M) means the number of parameters (million). }
  \resizebox*{\linewidth}{!}{
    \begin{tabular}{llccccccccc}
    \toprule
    \multirow{2}[4]{*}{} & \multirow{2}[4]{*}{Method} & \multicolumn{3}{c}{CIFAR10} & \multicolumn{3}{c}{CIFAR100} & \multicolumn{3}{c}{STL10} \\
\cmidrule{3-11}          &       & \#P(M) & top1 acc. & FPS   & \#P(M) & top1 acc. & FPS   & \#P(M) & top1 acc. & FPS \\
    \midrule
    \multirow{10}[2]{*}{\begin{sideways}ResNet83\end{sideways}} & SENet~\cite{hu2018squeeze} & 0.97  & 94.21  & 5277  & 0.99  & 74.62 & 5223  & 0.97  & 82.08  & 718  \\
          & LSAS-SENet & 0.95  & 94.32  ({\color{red}{$\uparrow$ 0.11}}) & 6192  ({\color{red}{$\uparrow$ 915}}) & 0.97  & 74.64 ({\color{red}{$\uparrow$ 0.02}}) & 6151  ({\color{red}{$\uparrow$ 928}}) & 0.95  & 84.89  ({\color{red}{$\uparrow$ 2.81}}) & 824  ({\color{red}{$\uparrow$ 106}}) \\
          & CBAM~\cite{woo2018cbam} & 0.97  & 93.31  & 2720  & 0.99  & 73.14  & 2692  & 0.97  & 81.81  & 428  \\
          & LSAS-CBAM & 0.95  & 93.51  ({\color{red}{$\uparrow$ 0.20}}) & 4547  ({\color{red}{$\uparrow$ 1827}}) & 0.97  & 73.53  ({\color{red}{$\uparrow$ 0.39}}) & 4509  ({\color{red}{$\uparrow$ 1817}}) & 0.95  & 82.33  ({\color{red}{$\uparrow$ 0.52}}) & 685  ({\color{red}{$\uparrow$ 257}}) \\
          & SRM~\cite{lee2019srm} & 0.89  & 94.55  & 4954  & 0.91  & 74.49 & 4929  & 0.89  & 81.44  & 664  \\
          & LSAS-SRM & 0.88  & 94.70  ({\color{red}{$\uparrow$ 0.15}}) & 6018  ({\color{red}{$\uparrow$ 1064}}) & 0.91  & 74.65 ({\color{red}{$\uparrow$ 0.16}}) & 5993  ({\color{red}{$\uparrow$ 1064}}) & 0.88  & 86.51  ({\color{red}{$\uparrow$ 5.07}}) & 812  ({\color{red}{$\uparrow$ 148}}) \\
          & ECA~\cite{2020ECA} & 0.87  & 93.98  & 5470  & 0.89  & 74.06  & 5456  & 0.87  & 81.34  & 720  \\
          & LSAS-ECA & 0.87  & 94.34  ({\color{red}{$\uparrow$ 0.36}}) & 6254  ({\color{red}{$\uparrow$ 784}}) & 0.90  & 74.4 ({\color{red}{$\uparrow$ 0.34}}) & 6233  ({\color{red}{$\uparrow$ 777}}) & 0.87  & 85.21  ({\color{red}{$\uparrow$ 3.87}}) & 824  ({\color{red}{$\uparrow$ 104}}) \\
          & SPANet~\cite{guo2020spanet} & 1.93  & 94.15 & 3788  & 1.96  & 74.64  & 3773  & 1.93  & 77.54  & 580  \\
          & LSAS-SPANet & 1.69  & 94.41 ({\color{red}{$\uparrow$ 0.26}}) & 4997  ({\color{red}{$\uparrow$ 1209}}) & 1.71  & 73.84  ({\color{green}{$\downarrow$ -0.80}}) & 4979  ({\color{red}{$\uparrow$ 1206}}) & 1.69  & 79.10  ({\color{red}{$\uparrow$ 1.56}}) & 724  ({\color{red}{$\uparrow$ 144}}) \\
    \midrule
    \multirow{10}[2]{*}{\begin{sideways}ResNet164\end{sideways}} & SENet~\cite{hu2018squeeze} & 1.91  & 94.57  & 2723  & 1.93  & 75.30  & 2715  & 1.91  & 83.81  & 368  \\
          & LSAS-SENet & 1.87  & 95.01  ({\color{red}{$\uparrow$ 0.44}}) & 3148  ({\color{red}{$\uparrow$ 425}}) & 1.89  & 76.47  ({\color{red}{$\uparrow$ 1.17}}) & 3140  ({\color{red}{$\uparrow$ 425}}) & 1.87  & 85.71  ({\color{red}{$\uparrow$ 1.90}}) & 423  ({\color{red}{$\uparrow$ 55}}) \\
          & CBAM~\cite{woo2018cbam} & 1.90  & 93.34  & 1364  & 1.93  & 73.25 & 1356  & 1.90  & 82.43 & 217  \\
          & LSAS-CBAM & 1.86  & 93.85  ({\color{red}{$\uparrow$ 0.51}}) & 2292  ({\color{red}{$\uparrow$ 928}}) & 1.89  & 74.03 ({\color{red}{$\uparrow$ 0.78}}) & 2282  ({\color{red}{$\uparrow$ 926}}) & 1.86  & 82.28  ({\color{green}{$\downarrow$ -0.15}}) & 351  ({\color{red}{$\uparrow$ 134}}) \\
          & SRM~\cite{lee2019srm} & 1.74  & 94.51  & 2510  & 1.76  & 74.56  & 2507  & 1.74  & 80.60  & 340  \\
          & LSAS-SRM & 1.73  & 94.63  ({\color{red}{$\uparrow$ 0.12}}) & 3053  ({\color{red}{$\uparrow$ 543}}) & 1.75  & 75.28  ({\color{red}{$\uparrow$ 0.72}}) & 3056  ({\color{red}{$\uparrow$ 549}}) & 1.73  & 85.09  ({\color{red}{$\uparrow$ 4.49}}) & 418  ({\color{red}{$\uparrow$ 78}}) \\
          & ECA~\cite{2020ECA} & 1.70  & 94.26  & 2781  & 1.73  & 74.49  & 2780  & 1.70  & 81.05  & 369  \\
          & LSAS-ECA & 1.71  & 94.63  ({\color{red}{$\uparrow$ 0.37}}) & 3184  ({\color{red}{$\uparrow$ 403}}) & 1.74  & 75.19  ({\color{red}{$\uparrow$ 0.70}}) & 3181  ({\color{red}{$\uparrow$ 401}}) & 1.71  & 84.85  ({\color{red}{$\uparrow$ 3.80}}) & 424  ({\color{red}{$\uparrow$ 55}}) \\
          & SPANet~\cite{guo2020spanet} & 3.83  & 94.31  & 1910  & 3.86  & 75.68  & 1910  & 3.83  & 75.33 & 296  \\
          & LSAS-SPANet & 3.34  & 94.71  ({\color{red}{$\uparrow$ 0.40}}) & 2531  ({\color{red}{$\uparrow$ 621}}) & 3.36  & 75.11  ({\color{green}{$\downarrow$ -0.57}}) & 2531  ({\color{red}{$\uparrow$ 621}}) & 3.34  & 79.55 ({\color{red}{$\uparrow$ 4.22}}) & 371  ({\color{red}{$\uparrow$ 75}}) \\
    \midrule
    \multirow{10}[2]{*}{\begin{sideways}ResNet245\end{sideways}} & SENet~\cite{hu2018squeeze} & 3.78  & 94.56 & 1366  & 3.80  & 75.95 & 1365  & 3.78  & 85.59  & 186  \\
          & LSAS-SENet & 3.70  & 94.96 ({\color{red}{$\uparrow$ 0.40}}) & 1577  ({\color{red}{$\uparrow$ 211}}) & 3.72  & 76.63 ({\color{red}{$\uparrow$ 0.68}}) & 1577  ({\color{red}{$\uparrow$ 212}}) & 3.70  & 86.24  ({\color{red}{$\uparrow$ 0.65}}) & 215  ({\color{red}{$\uparrow$ 29}}) \\
          & CBAM~\cite{woo2018cbam} & 3.77  & 93.4  & 682   & 3.79  & 71.23 & 680   & 3.77  & 80.18  & 109  \\
          & LSAS-CBAM & 3.69  & 94.16 ({\color{red}{$\uparrow$ 0.76}}) & 1146  ({\color{red}{$\uparrow$ 464}}) & 3.71  & 75.12 ({\color{red}{$\uparrow$ 3.89}}) & 1145  ({\color{red}{$\uparrow$ 465}}) & 3.69  & 81.54  ({\color{red}{$\uparrow$ 1.36}}) & 177  ({\color{red}{$\uparrow$ 68}}) \\
          & SRM~\cite{lee2019srm} & 3.44  & 93.43 & 1260  & 3.46  & 75.36 & 1259  & 3.44  & 74.35 & 172  \\
          & LSAS-SRM & 3.43  & 94.43 ({\color{red}{$\uparrow$ 1.00}}) & 1534  ({\color{red}{$\uparrow$ 274}}) & 3.45  & 76.06 ({\color{red}{$\uparrow$ 0.70}}) & 1536  ({\color{red}{$\uparrow$ 277}}) & 3.43  & 85.24 ({\color{red}{$\uparrow$ 10.89}}) & 212  ({\color{red}{$\uparrow$ 40}}) \\
          & ECA~\cite{2020ECA} & 3.37  & 94.22 & 1398  & 3.40  & 74.3  & 1396  & 3.37  & 78.45  & 187  \\
          & LSAS-ECA & 3.39  & 95.05 ({\color{red}{$\uparrow$ 0.83}}) & 1600  ({\color{red}{$\uparrow$ 202}}) & 3.42  & 75.43 ({\color{red}{$\uparrow$ 1.13}}) & 1599  ({\color{red}{$\uparrow$ 203}}) & 3.39  & 85.21  ({\color{red}{$\uparrow$ 6.76}}) & 215  ({\color{red}{$\uparrow$ 28}}) \\
          & SPANet~\cite{guo2020spanet} & 7.63  & 94.57 & 960   & 7.65  & 76.33  & 960   & 7.63  & 78.53  & 149  \\
          & LSAS-SPANet & 6.64  & 94.61 ({\color{red}{$\uparrow$ 0.04}}) & 1270  ({\color{red}{$\uparrow$ 310}}) & 6.66  & 75.86  ({\color{green}{$\downarrow$ -0.47}}) & 1270  ({\color{red}{$\uparrow$ 310}}) & 6.64  & 80.99  ({\color{red}{$\uparrow$ 2.46}}) & 188 ({\color{red}{$\uparrow$ 39}}) \\
    \bottomrule
    \end{tabular}%
    }
  \label{tab:main1}%
\end{table*}%

However, although multi-order sub-attention modules may mitigate the attention bias of the original self-attention modules, the learnable parameters $\gamma_i$ and $\beta_i$ inevitably increase the burden of DNNs on parameters and computation. 
In order to circumvent this burden, we design the selection gate as shown in Fig.~\ref{fig:LSAS} based on the past works~\cite{huang2022lottery} which reveals that it is not necessary to set a self-attention module for each block in the backbone~\cite{huang2020efficient} and the modules from the later blocks have a greater impact on model performance~\cite{hu2019squeeze,huang2020dianet}. 
Our selection gate makes DNNs only insert self-attention modules to the tail networks, whose structure is as follows:
\begin{equation}
\begin{aligned}
\mathbf{SG}(v) &= \begin{cases}                   
v, & \text{if } c(v) > \mu  \\
1, & \text{others},  
\end{cases}
\end{aligned}
\label{eq:select}
\end{equation}
where $c(\cdot)$ is the channel count function, which returns the channel number of input $v$. $\mu$ is the gate threshold for deciding to open and close the selection gate. The computation process of the attention module enhanced by LSAS can be referred to Algorithm~\ref{alg:ean}.

\section{Experiments}
\label{sec:exp}

In this section, we elaborate on the details of our experiments and investigate the effectiveness of our method on the image classification task.


We evaluate our method on four popular datasets, i.e., ImageNet~\cite{russakovsky2015imagenet}, STL10~\cite{coates2011analysis}, CIFAR10~\cite{krizhevsky2009learning} and CIFAR100~\cite{krizhevsky2009learning}. ImageNet has 1.28 million training images and 50k validation images of size 224 by 224 and has 1000 classes. STL10 has 5k train images and 8k test images of size 96 by 96 and has 10 classes. 
CIFAR10 and CIFAR100 have 50k train images and 10k test images of size 32 by 32 but have 10 and 100 classes respectively. We evaluate our method on ResNet~\cite{he2016deep} and several popular attention modules, including SENet~\cite{hu2018squeeze}, CBAM~\cite{woo2018cbam}, SRM~\cite{lee2019srm}, ECA~\cite{2020ECA}, and SPANet~\cite{guo2020spanet}.
We train all models on STL10, CIFAR10, and CIFAR100 with an Nvidia RTX 3080 GPU and set the epoch number to 164, and train the models on ImageNet with eight Nvidia RTX 3080 GPUs and set the epoch number to 100. SGD optimizer with a momentum of 0.9 and weight decay of $10^{-4}$ is applied. Furthermore, we use normalization and standard data augmentation, including random cropping and horizontal flipping during training. 


We compare the top 1 accuracy of different self-attention modules before and after using LSAS. 
The order $n$ of LSAS is uniformly set to 1 and further discussions are shown in Section \ref{sec:as}. The threshold $\mu$ in the selection gate is set to 512 on ImageNet and 128 on other datasets.
The experimental results shown in Table~\ref{tab:main1} clarify that LSAS improves most of the attention modules on different datasets and different network depths. 
For the small dataset CIFAR10, the performance of self-attention modules is good enough, so LSAS improves slightly.
But for CIFAR100 and STL10, most of the attention modules enhanced by LSAS have significant test accuracy improvement over the original attention modules.
Especially SRM performance becomes worse while LSAS-SRM shows stable performance improvement with increasing depth, which shows that LSAS improves the stability of attention modules while improving model accuracy.

\begin{table}[htbp]
  \centering
  \caption{Testing accuracy (\%) and Frames Per Second (FPS) on ImageNet. \#P(M) means the number of parameters (million). }
  \resizebox*{\linewidth}{!}{
    \begin{tabular}{clccc}
    \toprule
          & Method & \#P(M) & top1 acc. & FPS \\
    \midrule
    \multirow{4}[2]{*}{\begin{sideways}ResNet34\end{sideways}} 
          & SENet & 21.96 & 74.26  & 1588 \\
          & LSAS-SENet & 21.95 & 74.29 ({\color{red}{$\uparrow$ 0.03}}) & 1638 ({\color{red}{$\uparrow$ 50}}) \\
          & CBAM  & 21.96 & 74.01  & 1248 \\
          & LSAS-CBAM & 21.96 &  73.94({\color{green}{$\downarrow$ 0.07}})     & 1399 ({\color{red}{$\uparrow$ 151}}) \\
    \midrule
    \multirow{4}[2]{*}{\begin{sideways}ResNet50\end{sideways}} 
          & SENet & 28.09 & 76.63  & 772 \\
          & LSAS-SENet & 27.95 & 77.28 ({\color{red}{$\uparrow$ 0.65}})  & 827 ({\color{red}{$\uparrow$ 55}})\\
          & CBAM  & 28.09 & 76.40  & 547 \\
          & LSAS-CBAM & 27.95 & 76.75 ({\color{red}{$\uparrow$ 0.35}})  & 630 ({\color{red}{$\uparrow$ 83}}) \\
    \bottomrule
    \end{tabular}%
  }
  \label{tab:main2}%
\end{table}%

We analyze the complexity of LSAS in terms of the number of parameters and Frames Per Second (FPS). 
Compared to most baselines, LSAS does not result in a parameter increase due to the selection gate.
Even though LSAS results in a slight increase in the parameters of ECA, past works~\cite{huang2022layer,zhong2022mix} reveal that ECA performance is poor on these three datasets, while LSAS can greatly improve ECA performance which is more pronounced on large datasets.
FPS shown in Table~\ref{tab:main1} illustrates that LSAS also has significant advantages in computational efficiency. 

Within the multi-category and high-resolution image, LSAS also has superior and stable performance improvement as shown in Table~\ref{tab:main2}.
In summary, LSAS improves the performance of self-attention modules while reducing the number of parameters and increasing the computation speed of DNNs.

\section{Ablation study}
\label{sec:as}
In this section, we use ResNet164 to analyze LSAS in terms of the order of sub-attention and selection gate on STL10.

\subsection{The Order of Sub-attention}
We set the order $n$ from 0 to 5, and the experimental results are shown in Table~\ref{tab:order}. 
For LSAS-SENet and LSAS-CBAM, the performance is best when $n$ is 2 and 1, respectively. And the performance of LSAS-SENet is second best while $n=1$.
Therefore, for a given self-attention module, we recommend setting $n$ to 1. Moreover, there are at least two other reason for $n=1$. 
On the one hand, the smaller $n$, the smaller the impact of LSAS on the number of parameters and computation of DNNs. 
On the other hand, since the essence of multi-order sub-attention is multiplication $\Pi^n_iv_i^{'}$ and the value of $v_i^{'} \leq 1$, the increase in $n$ means accumulation increasing which results in a smaller product result. 
If $n$ is too large, $v_0{'}$ will be too small and affect the information forward of the backbone network, leading to poor prediction. 
This inference is consistent with the experimental results of Table~\ref{tab:order}.

\begin{table}[htbp]
  \centering
  \caption{Testing accuracy (\%) of ResNet164 with different attention modules and order $n$ on STL10. \#P(M) means the number of parameters (million). Bold and underline indicate the best results and the second best results, respectively.}
  \resizebox*{0.75\linewidth}{!}{
    \begin{tabular}{ccccc}
    \toprule
    \multirow{2}[4]{*}{$n$} & \multicolumn{2}{c}{LSAS-SENet} & \multicolumn{2}{c}{LSAS-CBAM} \\
\cmidrule{2-5}          & \#P(M) & top1 acc. & \#P(M) & top1 acc. \\
    \midrule
    0     & 1.86  & 80.24 & 1.85  & \underline{81.19} \\
    1     & 1.87  & \underline{85.71} & 1.86  & \textbf{82.28} \\
    2     & 1.87  & \textbf{85.93} & 1.87  & 80.15 \\
    3     & 1.88  & 85.34 & 1.88  & 79.41 \\
    4     & 1.89  & 82.20 & 1.89  & 78.75 \\
    5     & 1.90  & 81.86 & 1.90  & 77.48 \\
    \bottomrule
    \end{tabular}%
  }
  \label{tab:order}%
\end{table}%

\subsection{Gate Threshold of Selection Gate}


Gate threshold $\mu$ determines the number of blocks enhanced by attention modules.
The larger $\mu$, the fewer the number of blocks with attention. 
According to the structure of ResNet, we explore the performance of ResNet164 when $\mu$ is 0, 64, 128, and 256. 
As shown in Table~\ref{tab:gate}, when the value of $\mu$ is 64, LSAS-SENet achieves the best performance, and when $\mu$ is 128 the performance of DNNs is close to the optimal performance. These indicate that more attention modules don't mean better performance, which is consistent with the idea of the selection gate. 
By comparing to Table~\ref{tab:main1}, LSAS has an advantage in parameter quantity when $\mu$ is 128 or 256. However, when $\mu$ is 256, the model degenerates into vanilla ResNet, the accuracy of which is far from optimal.
Therefore, we recommend setting $\mu$ to 128. 

\begin{table}[htbp]
  \centering
  \caption{Testing accuracy (\%) and Frames Per Second (FPS) of ResNet164 with different attention modules and gate threshold $\mu$ on STL10. \#P(M) means the number of parameters (million). }
  \resizebox*{\linewidth}{!}{
    \begin{tabular}{ccccccc}
    \toprule
    \multirow{2}[4]{*}{$\mu$} & \multicolumn{3}{c}{LSAS-SENet} & \multicolumn{3}{c}{LSAS-CBAM} \\
\cmidrule{2-7}          & \#P(M) & top1 acc. & FPS   & \#P(M) & top1 acc. & FPS \\
    \midrule
    0     & 1.93  & 85.94 & 369   & 1.92  & 82.96 & 172 \\
    64    & 1.92  & 86.26 & 405   & 1.91  & 82.80 & 265 \\
    128   & 1.87  & 85.71 & 426   & 1.86  & 82.28 & 358 \\
    256   & 1.70  & 80.94 & 438   & 1.70  & 80.94 & 438 \\
    \bottomrule
    \end{tabular}%
  }
  \label{tab:gate}%
\end{table}%

\section{Conclusion}

We focus on the attention bias problem in DNNs and show the quantitative analysis of the attention bias of DNNs by proposing AE and AE-datasets.
By using AE, we find that the self-attention mechanism can alleviate the attention bias problem of DNNs, while the attention mechanism still has a non-negligible bias problem.
Inspired by the attention mechanism, we propose LSAS to further alleviate the attention bias problem of the attention mechanism. 
LSAS comprises multi-order sub-attention and a selection gate strategy.
The selection gate determines the embedding of the multi-level sub-attention module at the end of DNNs, which realizes controlling the number of parameters and computation.
Experiments on multiple datasets and multiple attention modules show that LSAS can effectively enhance the attention modules and improve the attention ability and computational efficiency of DNNs while reducing the amount of DNN parameters.

\bibliographystyle{IEEEbib}
\bibliography{refs}

\begin{thebibliography}{10}

\bibitem{liang2020instance}
Senwei Liang, Zhongzhan Huang, Mingfu Liang, and Haizhao Yang,
\newblock ``Instance enhancement batch normalization: An adaptive regulator of
  batch noise,''
\newblock in {\em Proceedings of the AAAI Conference on Artificial
  Intelligence}, 2020, vol.~34, pp. 4819--4827.

\bibitem{kuang2021face}
Qing Kuang,
\newblock ``Face image feature extraction based on deep learning algorithm,''
\newblock in {\em Journal of Physics: Conference Series}. IOP Publishing, 2021,
  vol. 1852, p. 032040.

\bibitem{he2016deep}
Kaiming He, Xiangyu Zhang, Shaoqing Ren, and Jian Sun,
\newblock ``Deep residual learning for image recognition,''
\newblock in {\em Proceedings of the IEEE conference on computer vision and
  pattern recognition}, 2016, pp. 770--778.

\bibitem{zeiler2014visualizing}
Matthew~D Zeiler and Rob Fergus,
\newblock ``Visualizing and understanding convolutional networks,''
\newblock in {\em European conference on computer vision}. Springer, 2014, pp.
  818--833.

\bibitem{mcnamara2017developing}
Quinten McNamara, Alejandro De~La~Vega, and Tal Yarkoni,
\newblock ``Developing a comprehensive framework for multimodal feature
  extraction,''
\newblock in {\em Proceedings of the 23rd ACM SIGKDD International Conference
  on Knowledge Discovery and Data Mining}, 2017, pp. 1567--1574.

\bibitem{zhu2021residual}
Ke~Zhu and Jianxin Wu,
\newblock ``Residual attention: A simple but effective method for multi-label
  recognition,''
\newblock in {\em Proceedings of the IEEE/CVF International Conference on
  Computer Vision}, 2021, pp. 184--193.

\bibitem{guo2021ssan}
Xudong Guo, Xun Guo, and Yan Lu,
\newblock ``Ssan: Separable self-attention network for video representation
  learning,''
\newblock in {\em Proceedings of the IEEE/CVF Conference on Computer Vision and
  Pattern Recognition}, 2021, pp. 12618--12627.

\bibitem{2017Grad}
Ramprasaath~R. Selvaraju, Michael Cogswell, Abhishek Das, Ramakrishna Vedantam,
  Devi Parikh, and Dhruv Batra,
\newblock ``Grad-cam: Visual explanations from deep networks via gradient-based
  localization,''
\newblock in {\em International Conference on Computer Vision}, 2017.

\bibitem{huang2020dianet}
Zhongzhan Huang, Senwei Liang, Mingfu Liang, and Haizhao Yang,
\newblock ``Dianet: Dense-and-implicit attention network.,''
\newblock in {\em AAAI}, 2020, pp. 4206--4214.

\bibitem{hu2018squeeze}
Jie Hu, Li~Shen, and Gang Sun,
\newblock ``Squeeze-and-excitation networks,''
\newblock in {\em Proceedings of the IEEE conference on computer vision and
  pattern recognition}, 2018, pp. 7132--7141.

\bibitem{anderson2005cognitive}
John~R Anderson,
\newblock {\em Cognitive psychology and its implications},
\newblock Macmillan, 2005.

\bibitem{woo2018cbam}
Sanghyun Woo, Jongchan Park, Joon-Young Lee, and In~So Kweon,
\newblock ``Cbam: Convolutional block attention module,''
\newblock in {\em Proceedings of the European conference on computer vision
  (ECCV)}, 2018, pp. 3--19.

\bibitem{lee2019srm}
HyunJae Lee, Hyo-Eun Kim, and Hyeonseob Nam,
\newblock ``Srm: A style-based recalibration module for convolutional neural
  networks,''
\newblock in {\em Proceedings of the IEEE/CVF International Conference on
  Computer Vision}, 2019, pp. 1854--1862.

\bibitem{2020ECA}
Q.~Wang, B.~Wu, P.~Zhu, P.~Li, and Q.~Hu,
\newblock ``Eca-net: Efficient channel attention for deep convolutional neural
  networks,''
\newblock in {\em 2020 IEEE/CVF Conference on Computer Vision and Pattern
  Recognition (CVPR)}, 2020.

\bibitem{guo2020spanet}
Jingda Guo, Xu~Ma, Andrew Sansom, Mara McGuire, Andrew Kalaani, Qi~Chen, Sihai
  Tang, Qing Yang, and Song Fu,
\newblock ``Spanet: Spatial pyramid attention network for enhanced image
  recognition,''
\newblock in {\em 2020 IEEE International Conference on Multimedia and Expo
  (ICME)}. IEEE, 2020, pp. 1--6.

\bibitem{huang2022lottery}
Zhongzhan Huang, Senwei Liang, Mingfu Liang, Wei He, Haizhao Yang, and Liang
  Lin,
\newblock ``The lottery ticket hypothesis for self-attention in convolutional
  neural network,''
\newblock {\em arXiv preprint arXiv:2207.07858}, 2022.

\bibitem{huang2020efficient}
Zhongzhan Huang, Senwei Liang, Mingfu Liang, Wei He, and Haizhao Yang,
\newblock ``Efficient attention network: Accelerate attention by searching
  where to plug,''
\newblock {\em arXiv preprint arXiv:2011.14058}, 2020.

\bibitem{hu2019squeeze}
Jie Hu, Li~Shen, Samuel Albanie, Gang Sun, and Enhua Wu,
\newblock ``Squeeze-and-excitation networks,''
\newblock {\em IEEE Transactions on Pattern Analysis and Machine Intelligence},
  vol. 42, no. 8, pp. 2011--2023, 2019.

\bibitem{russakovsky2015imagenet}
Olga Russakovsky, Jia Deng, Hao Su, Jonathan Krause, Sanjeev Satheesh, Sean Ma,
  Zhiheng Huang, Andrej Karpathy, Aditya Khosla, Michael Bernstein, et~al.,
\newblock ``Imagenet large scale visual recognition challenge,''
\newblock {\em International journal of computer vision}, vol. 115, no. 3, pp.
  211--252, 2015.

\bibitem{coates2011analysis}
Adam Coates, Andrew Ng, and Honglak Lee,
\newblock ``An analysis of single-layer networks in unsupervised feature
  learning,''
\newblock in {\em Proceedings of the fourteenth international conference on
  artificial intelligence and statistics}. JMLR Workshop and Conference
  Proceedings, 2011, pp. 215--223.

\bibitem{krizhevsky2009learning}
Alex Krizhevsky, Geoffrey Hinton, et~al.,
\newblock ``Learning multiple layers of features from tiny images,''
\newblock 2009.

\bibitem{huang2022layer}
Zhongzhan Huang, Senwei Liang, Mingfu Liang, Weiling He, and Liang Lin,
\newblock ``Layer-wise shared attention network on dynamical system
  perspective,''
\newblock {\em arXiv preprint arXiv:2210.16101}, 2022.

\bibitem{zhong2022mix}
Shanshan Zhong, Wushao Wen, and Jinghui Qin,
\newblock ``Mix-pooling strategy for attention mechanism,''
\newblock {\em arXiv preprint arXiv:2208.10322}, 2022.

\end{thebibliography}

\end{document}